# Measuring Human-Robot Trust with the MDMT (Multi-Dimensional Measure of Trust)*

Bertram F. Malle and Daniel Ullman

*Abstract*— We describe the steps of developing the MDMT (Multi-Dimensional Measure of Trust), an intuitive self-report measure of perceived trustworthiness of various agents (human, robot, animal). We summarize the evidence that led to the original four-dimensional form (v1) and to the most recent five-dimensional form (v2). We examine the measure's strengths and limitations and point to further necessary validations.

## I. Background

A number of trust measures have been used in the human-robot interaction literature, but they suffer from one or more of the following limitations: they are time consuming to complete; they are not designed for human-robot relations but adapted from human-human relationships or from the automation literature; they lack a theoretical framework; they have psychometric shortcomings. In this context, we decided in 2017 to develop a measure of trust that tries to overcome some or all of these limitations. We aimed for a short, intuitive instrument that is suitable to assess people's trust in other humans, robots, AI, or even animals, is informed by theories of trust, and has demonstrable reliability and validity [1]–[4]. This is a summary on how we developed this measure— the Multi-Dimensional Measure of Trust (MDMT) [5], [6]—and where we must go next.

## II. MDMT v1

From the start, our aim was not to measure trust merely as an undifferentiated feeling state. A one-item measure ("Do you feel you can trust $X$?") may suffice here [7]–[10]. In many contexts, this question may be too vague, and we should ask, "Do you trust $X$ to do $Y$?" If $Y$ has been properly specified, a one-item measure may once more suffice. In other contexts, however, $Y$ cannot be specified, and we would need to capture the range of expectations that a "trustor" has about the "trustee's" various actions $Y_i$ that could endanger the trustor's vulnerability [3]. Such expectations are based on beliefs the trustor has about the trustee's abilities, intentions, and commitments. These are beliefs about the other's *trustworthiness*—relevant attributes that guide the trustee's actions $Y_i$. Attributes of trustworthiness are unlikely to be unidimensional—we may trust a person to perform well on a risky task because they are highly competent; we may trust another to tell us the truth because they are candid and sincere; and we may trust a third to benefit us, even if it slightly bends the rules. Thus, we framed our intended measure of trust as a multi-dimensional set of expectations about another agent's trustworthiness attributes, whether person, machine, etc.

We initially assumed that there were two broad dimensions of trust—capacity trust (studied in the automation literature) and moral-personal trust (studied in the psychological literature). We tested this hypothesis in [1] by selecting 62 words or phrases that referred to various attributes of trustworthiness (e.g., skilled, reliable, honest, responsible), collated from the scientific literature and a Thesaurus. We presented these words to participants and asked them to indicate where they thought each word falls on a scale from "more similar to capacity trust" (defined as "trusting that an agent is capable of completing a task) to "more similar to personal trust" (defined as "trusting that an agent will not place you at risk"). We analyzed the resulting covariance matrix with iterations of Principal Components Analysis (PCA), item analysis, and item selection. To our surprise, the ratings were best described by more than two dimensions—in fact, the data suggested a structure of four dimensions, best labeled *Capable* (e.g., diligent, rigorous,), *Reliable* (e.g., can depend on, can count on), *Ethical* (e.g., principled, scrupulous), and *Sincere* (e.g., genuine truthful). (For more details, see [1] and https://research.clps.brown.edu/SocCogSci/Measures.)

To further validate this emerging model, in [3] we presented participants with a list of 32 trust-relevant attributes (many from the initial study [1]), as well as 5 trust-unrelated filler items, and asked them to sort the terms into five bins. Four bins were introduced as "person types" each represented by a single character trait (Reliable, Capable, Sincere, and Ethical). A fifth bin was labeled "Other," for words that did not fit the first four. The results were quite clear (see [3], Fig. 1): The hypothesized bins contained largely the same items as in the initial study [1] and people showed agreement rates of 51% to 88% for the top items in each bin.

From the results of these initial studies we developed the MDMT v1 [5], with four attributes each representing the dimensions of Capable, Reliable, Ethical, and Sincere. In addition to rating the degree to which the agent had each attribute, participants had the option, for every item, to refrain from giving a specific rating and instead indicate that the particular term "Does Not Fit." Robots of lower complexity routinely elicit more DNF selections, especially for the Moral subscales [4], and allowing people to "opt out" of certain items improves interpretability of trust ratings [11].

## III. MDMT v2

Even though the initial studies suggested four distinct dimensions, our review of the literature had suggested that a fifth dimension may be important, at least in human-human trust but potentially also in human-machine trust: that of

* This work was supported in part by a grant from the Air Force Office of Scientific Research Award No. FA9550-21-1-0359.

Bertram F. Malle is with the Department of Cognitive, Linguistic, and Psychological Sciences, Brown University, Providence, RI, USA (bfmalle@brown.edu). Daniel Ullman is with Meta Reality Labs, New York, NY, USA.

Benevolence (see Table 2 in [3]). As part of his dissertation [4], Danny Ullman therefore conducted a new study, this time presenting 41 trust-related terms to participants and making an explicit attempt to include attributes that captured facets of benevolence (e.g., "kind," "looks out for others," "shows goodwill"). Participants were again asked to classify the terms into bins, but this time into any number of bins between two and six, and the bins were not labeled. After the sorting task, participants had a chance to label the bins that they had used.

The results of this study were compelling. First, 74% of participants sorted the trust-related terms into four, five, or six bins, even though using fewer bins would have been easier. Second, people's own labels for their bins arranged themselves in a consistent and systematic manner: the five most frequently used labels corresponded neatly to the four previous dimensions plus the suspected Benevolence dimension. The next five most frequently used labels showed the same pattern; and so did the next five (see Table 1).

TABLE I. PEOPLE'S OWN LABELS FOR BINS INTO WHICH THEY SORTED TRUST-RELATED TERMS

| Hypothesized dimension | Five most frequent labels | Next five most frequent labels | Next five most frequent labels |
|---|---|---|---|
| COMPETENT | Competent (29)[a] | Skilled (17) | Capable (14) |
| RELIABLE | Dependable (23) | Reliable (14) | Responsible (14) |
| ETHICAL | Moral (30) | Loyal (15) | Ethical (11) |
| TRANSPARENT | Honest (28) | Authentic (15) | Candid (13) |
| BENEVOLENT | Caring (21) | Kind (20) | Benevolent (13) |

a. Numbers in parentheses refer to how many people (out of 92) used that label for one of their bins

Most important, analyses of the co-occurrence matrix (how often each term shared the same bin with a given other term) supported a five-dimensional model, whether we used cluster analysis, network analysis, PCA, or MDS. Fig. 1 shows the result of the *k*-means cluster analysis projected into 2D space, which illustrates a Performance axis (Reliable, Competent) and a Moral axis (Ethical, Transparent, Benevolent).

In light of these results we revised the MDMT to version 2 [6], which contains four items for each of five dimension subscales and thus 20 items total.

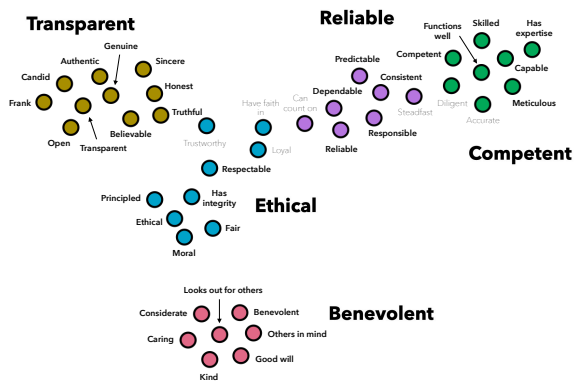

Figure 1. *k*-means cluster analysis of 41 trust-related terms that 92 participants sorted into a freely chosen number of "groups you feel belong together"

IV. STRENGTHS, LIMITATIONS, AND NEXT STEPS

One strength of the MDMT is its flexibility [13]—specific subscales can be selected when most suitable for the "trustee" in question. For example, trust in automation (e.g., self-driving cars) may get by with the Performance-related subscales, but trust in a language-capable social robot may require including the Morality-related subscales. Importantly, whenever the researcher asks participants about dimensions they find absurd to consider (e.g., the ethical integrity of a convolutional net), the "Does Not Fit" option is a welcome exist strategy

Though the 20-item-MDMT v2 is a compact measure (each item consists of only 1-3 words), it still takes a few minutes to complete. For studies in which one would like to assess people's changing trust over time, we have recently found that 10-item short forms (2 items per subscale) work well and still show good psychometric properties. This allows researchers to measure trust updates over time [7] without allowing participants to slip into a mindless response set of filling out the same 20 items each and every time.

The MDMT v1 and v2 have shown excellent internal consistency (α values in the .70s to .90s), both at the factor level (Performance, Moral) and the dimension subscale level (Reliable, Competent, etc.). The correlations among subscales, however, tend to be high (*r*s often above .50), and PCA results typically point to only two factors [2], [12]. It is possible that people's conceptual space of trust (as evidenced in the sorting study [4]) is more differentiated than their actual judgments of a given agent (whether robot or human). However, it is also possible that such judgments sometimes do differentiate, even when they are often highly correlated. PCA is designed to clump correlated items together, so it may not be the best way to justify separation of correlated subscales; tests of predictive validity may be more useful.

For example, in the final study of Danny Ullman's dissertation [4], all five subscales systematically and differentially changed in response to dimension-specific new information about an agent (robot or human). For example, the Benevolence subscale changed more strongly to information about kindness than the other subscales did; and the Benevolence subscale also responded more strongly to kindness information than it did to any other information (e.g., about reliability, sincerity). (See [7] for a similar case of differential scale responses to experimental manipulations.)

Future validation research needs to demonstrate that the MDMT subscales are indeed differentially responsive in specific contexts, even when they are highly convergent in other contexts. The challenge is that new measures have to be validated against well-understood, standard criteria (stimuli, manipulations, etc.), but such criteria are largely absent in HRI or have themselves not yet been validated. Thus, we must engage in a bootstrapping process that helps us build valid criteria at the same time as we improve the measures, like the MDMT, that should be sensitive to these criteria.

We hope that the research community will continue to find use in the MDMT and that, as a community, we can refine it and determine the scope and limits of its validity. This workshop should allow us to take further steps in this direction.